\documentclass[letterpaper, 10 pt, conference]{ieeeconf}

\IEEEoverridecommandlockouts
\usepackage[margin=0.75in]{geometry}

\usepackage[utf8]{inputenc}

\let\labelindent\relax

\usepackage{amsmath}
\usepackage{amsthm}
\usepackage{amssymb}
\usepackage{enumitem} 
\usepackage{nicefrac}       \usepackage{microtype}      \usepackage{hyperref}

\usepackage[noadjust]{cite}

\usepackage{pifont}

\usepackage{url}

\usepackage{xspace}
\usepackage{xcolor}
\usepackage{booktabs} \usepackage{tikz}
\usetikzlibrary{automata}
\usetikzlibrary{positioning}
\usetikzlibrary{patterns}
\usetikzlibrary{shapes,arrows}
\tikzset{state/.style={circle, draw, minimum size=0.5cm}}

\usepackage{algorithm}
\usepackage{algorithmicx}
\usepackage{algpseudocode}
\algnewcommand{\LineComment}[1]{\Statex \(\triangleright\) #1}
\makeatletter
\newcommand{\algmargin}{\the\ALG@thistlm}
\makeatother
\newlength{\whilewidth}
\algdef{SE}[parWHILE]{parWhile}{EndparWhile}[1]
  {\parbox[t]{\dimexpr\linewidth-\algmargin}{     \hangindent\whilewidth\strut\algorithmicwhile\ #1\ \algorithmicdo\strut}}{\algorithmicend\ \algorithmicwhile}\algnewcommand{\parState}[1]{\State  \parbox[t]{\dimexpr\linewidth-\algmargin}{\strut #1\strut}}
\algnewcommand{\parRequire}[1]{\Require  \parbox[t]{\dimexpr\linewidth-\algmargin}{\strut #1\strut}}

\usepackage{cleveref}

\newtheorem{definition}{Definition}

\newtheorem{lemma}{Lemma}

\crefname{section}{Section}{Sections}
\crefname{subsection}{Section}{Sections}
\crefname{definition}{Definition}{Definitions}
\crefname{proposition}{Proposition}{Propositions}
\crefname{lemma}{Lemma}{Lemmas}
\crefname{theorem}{Theorem}{Theorems}
\crefname{corollary}{Corollary}{Corollaries}
\crefname{example}{Example}{Examples}
\crefname{figure}{Figure}{Figures}
\crefname{table}{Table}{Tables}
\crefname{assumption}{Assumption}{Assumptions}
\crefname{remark}{Remark}{Remarks}
\crefname{running}{Running Example}{Running Examples}
\crefname{algorithm}{Algorithm}{Algorithms}

\usepackage{amsmath}
\usepackage{amssymb}
\usepackage{xspace}
\usepackage{latexsym}
\usepackage{mathtools}
\usepackage{relsize}

\newcommand{\nat}{\mathbb{N}}

\newcommand{\real}{\mathbb{R}}

\newcommand{\abs}[1]{\vert #1 \vert}

\newcommand{\ap}{\mathsf{a}} \newcommand{\aps}{\mathsf{AP}} \newcommand{\pv}{\pi} \newcommand{\pvs}{\Pi} \newcommand{\sig}{\sigma}  

\newcommand{\G}{\Box}
\newcommand{\F}{\Diamond}
\newcommand{\X}{\bigcirc}
\newcommand{\U}{\mathbin{\mathcal{U}}}

\newcommand{\True}{\ensuremath{\mathtt{T}}}

\newcommand{\m}{\mathcal{M}}
\newcommand{\s}{\mathcal{A}}
\newcommand{\mss}{\mathtt{S}} 
\newcommand{\ms}{\mathtt{s}}
\newcommand{\msi}{\mathtt{s_0}}

\newcommand{\mas}{\mathtt{A}}
\newcommand{\ma}{\mathtt{a}}

\newcommand{\mts}{\mathtt{T}}
\newcommand{\ml}{\mathtt{L}}

\newcommand{\str}{\mathrm{str}}

\usepackage{todonotes}

\title{\bf Hyperproperties for Robotics: Planning via HyperLTL}

\author{Yu Wang, Siddhartha Nalluri, and Miroslav Pajic\thanks{This work is sponsored in part by the ONR under agreements N00014-17-1-2504 and N00014-20-1-2745, AFOSR under award number FA9550-19-1-0169, as well as the NSF CNS-1652544 award.}\thanks{Yu Wang and Miroslav Pajic are with the Department of Electrical and Computer Engineering, Duke University, Durham, NC 27708,~USA, {\tt\small \{yu.wang094, miroslav.pajic\}@duke.edu}. Siddhartha Nalluri is with the Computer Science Department, Duke University, Durham, NC 27708, USA, {\tt\small siddhartha.nalluri@duke.edu}.}}

\begin{document}

\maketitle
\thispagestyle{empty}
\pagestyle{empty}

\begin{abstract}
There is a growing interest on formal methods-based robotic planning for temporal logic objectives. In this work, we extend the scope of existing synthesis methods to hyper-temporal logics. We are motivated by the fact that important planning objectives, such as optimality, robustness, and privacy, (maybe implicitly) involve the interrelation between multiple paths. Such objectives are thus hyperproperties, and cannot be expressed with usual temporal logics like the linear temporal logic (LTL). We show that such hyperproperties can be expressed by HyperLTL, an extension of LTL to multiple paths. To handle the complexity of planning with HyperLTL specifications, we introduce a symbolic approach for synthesizing planning strategies on discrete transition systems. Our planning method is evaluated on several case studies.
\end{abstract}

\section{Introduction} \label{sec:intro}

The past decade has seen an increasing interest on planning problems from temporal logic objectives (e.g.,~\cite{fainekos_TemporalLogicMotion_2005,plaku_MotionPlanningTemporallogic_2016,elfar_cav19}).
Using temporal logics, such as the linear temporal logic (LTL)~\cite{kress-gazit_TemporalLogicBasedReactiveMission_2009,moarref_ReactiveSynthesisRobotic_2018,kress-gazit_WhereWaldoSensorBased_2007}, 
a wide class of objectives beyond reachability can be defined; these include infinite recurrence, complex dependency of many tasks~\cite{he_ReactiveSynthesisFinite_2017}, 
and time-dependent formations of multiple robotics~\cite{kantaros_GlobalPlanningMultiRobot_2016,kantaros_TemporalLogicTask_2019,kantaros_STyLuSTemporalLogic_2018}.

However, temporal logics commonly used in robotics (e.g., for planning) can only specify properties for individual executions (i.e., paths).
This effectively prevents them~from capturing important planning objectives, such as optimality, robustness, and privacy/opacity, which involve inter-relations between multiple paths. 
For example, to {(directly) specify an optimal strategy for a temporal logic objective, we should} ask for the existence of a path $\pv$ such that all other paths $\pv'$ are no better than $\pv$.
{Previous works on optimal or robust planning with temporal logic objectives rely on converting these objectives into cost functions~\cite{farahani_RobustModelPredictive_2015,lindemann_RobustMotionPlanning_2017,wolff_OptimalControlNondeterministic_2013}, or into games~\cite{jing_ShortcutEvilDoor_2013}.
Such conversions are specific to individual problems.}

In this paper, we show that such objectives, which directly specify the relations of multiple paths (or computation trees) and are commonly referred to as \emph{hyperproperties}~\cite{clarkson_Hyperproperties_2008}, 
can be \emph{generally} and \emph{formally} specified by hyper-temporal logics, such as HyperLTL~\cite{clarkson_TemporalLogicsHyperproperties_2014}.\footnote{It is worth noting that non-hyper temporal logics, such as the Computation Tree Logic (CTL), cannot express such 
relations between multiple non-nested computation trees; HyperLTL subsumes CTL~\cite{clarkson_Hyperproperties_2008}.}
HyperLTL extends LTL with a set of path variables to denote individual paths, and associates each atomic proposition with a path variable to indicate on which path it should hold.
HyperLTL also allows for the use of ``exists'' $\exists$ and ``for all'' $\forall$ quantifications of the path variables; this enables specifying relevant planning objectives such as the described planning optimality requirement that employs $\exists \pv \forall \pv'$~quantifiers.

We also propose a symbolic method to 
synthesize planning strategies with performance (e.g., optimality, robustness, and privacy) guarantees specified by HyperLTL. Specifically, we study the planning from HyperLTL objectives on a commonly used modeling formalism -- discrete transition systems (DTSs), where the states are discrete and the transitions are driven by actions.
This model can be viewed as a high-level discrete abstraction of the full workspace~\cite{kantaros_GlobalPlanningMultiRobot_2016,kantaros_TemporalLogicTask_2019}, that is obtained by the low-level explorations, like RRT or probabilistic roadmaps~\cite{lavalle_PlanningAlgorithms_2006}, or from the abstraction and simulation~\cite{fainekos_TemporalLogicMotion_2005}.

For finite-state discrete models like finite-state DTS, feasible strategies for temporal logic objectives, such as LTL, can be synthesized using automata-theoretic model checking (e.g., \cite{fainekos_TemporalLogicMotion_2005,lu_FrameworkModelChecking_2014,schillinger_MultiobjectiveSearchOptimal_2017,kupferman_AutomataTheoryModel_2018}). 
Specifically, the objective is first converted to an automaton, and then the strategy-search is performed on the product of the automaton and the discrete system model. 
However, this approach is extremely computationally intensive for HyperLTL objectives,  
since the possible quantifier alternation, e.g., $\exists \pv \forall \pv'$ for optimality objectives, dramatically increases the state of corresponding automata.
In addition, to keep track of the $n$ paths involved in $\varphi$ (e.g., $n=2$ for 
the optimality objectives since they employ two path variables $\pv$ and $\pv'$, as we  introduce in~\eqref{eq:optimality_general}), the $n$-fold self-product of the discrete system model is needed.  
As a result, in the formal methods community, automata-theoretic model checking of HyperLTL is mainly confined to quantifier-alternation-free objectives~\cite{finkbeiner_AlgorithmsModelChecking_2015}.

Consequently, to mitigate the state explosion for HyperLTL objectives in robotic planning, we adopt a symbolic approach for synthesizing strategies via SMT solvers~\cite{biere_SATBasedModelChecking_2018,biere_BoundedModelChecking_2003}. 
Specifically, the dynamics of the~DTS model is converted into a set of logic formulas, and feasible strategies should satisfy the conjunction of the HyperLTL objectives and model dynamics.
This conjunction is a first-order logic formula whose solution can be obtained by using off-the-shelf SMT solvers such as Z3~\cite{demoura_Z3EfficientSMT_2008}, Yices~\cite{dutertre_YICESSMTSolver_2006} or CVC4~\cite{barrett_CVC4_2011}.
As with previous work on symbolic synthesis from regular LTL (e.g.,~\cite{biere_BoundedModelChecking_2003,he_EfficientSymbolicReactive_2019,he_ReactiveSynthesisFinite_2017,huang_ControllerSynthesisLinear_2016}), we focus on HyperLTL objectives with a bounded time horizon~$T$, which we referred to as HyperLTLf.

Compared to the automata-theoretic approach, the symbolic synthesis method yields a more compact representation of the regular planning models; thus, it reduces the synthesis complexity by avoiding constructing the $n$-fold self product~\cite{he_EfficientSymbolicReactive_2019,he_ReactiveSynthesisFinite_2017}, and can even handle planning on DTSs with infinite states~\cite{huang_ControllerSynthesisInductive_2015,huang_ControllerSynthesisLinear_2016}. 
We show in case studies that our symbolic synthesis approach effectively handles planning problems with hyper temporal logic objectives, deriving strategies with strong robustness, privacy/opacity, and optimality guarantees.

This paper is organized as follows. After preliminaries (\cref{sec:prelim}), we show the need for the use of hyperproperties in path planning (\cref{sec:motivation}).
We formulate our planning problem on DTSs in \cref{sec:setup}, and show how HyperLTL objectives should be used to ensure robustness, privacy/opacity, and optimality of derived plans (\cref{sec:logic}).
We introduce a symbolic synthesis method for the HyperLTL objectives on the DTSs in \cref{sec:method}, evaluate it on case studies in \cref{sec:cases}, and finally conclude this work in \cref{sec:conc}.

\section{Preliminaries} \label{sec:prelim}
\paragraph*{Notation} The sets of integers and real numbers are denoted by $\nat$ and $\real$.
The phrase ``if and only if'' is abbreviated as ``iff''.
For $n \in \nat$, let $\nat_\infty = \nat \cup \{\infty\}$ and $[n] = \{1,\ldots,n\}$.
The cardinality and the power set of a set are denoted by $\abs{\cdot}$ and $2^{\cdot}$, and $\emptyset$ denotes the empty set.

\paragraph*{Linear Temporal Logic (LTL)} \label{sub:ltl}

Let $\aps$ be a set of properties (atomic propositions) related to the planning objective.
Formally, an LTL objective (i.e., specification) is constructed inductively by (the syntax) 
\vspace{-4pt}
\[ \label{eq:ltl_syntax}
    \varphi \Coloneqq \ap 
    \ \vert \ \neg \varphi
    \ \vert \ \varphi \land \varphi
    \ \vert \ \varphi \X \varphi
    \ \vert \ \varphi \U_T \varphi ,\vspace{-4pt}
\]
where $T \in \nat_\infty$ and $\ap \in \aps$; note that using ``bounded until'' $\U_T$ instead of the common ``until'' $\U$ is equivalent in defining LTL~\cite{pnueli1977temporal}. 
For a path $\pv: \nat \rightarrow 2^\aps$, where $\pv(t)$ is the set of properties satisfied at time $t$ by $\pv$,  
satisfaction of an LTL formula on path $\pv$ is checked recursively via (the semantics)
\vspace{-4pt}
\[ \label{eq:ltl_semantic}
    \begin{array}{l@{\hspace{1em}}c@{\hspace{1em}}l}
    \pv \models \ap & \Leftrightarrow & \ap \in \pv (0)
    \\ \pv \models \neg \varphi & \Leftrightarrow & \pv \not\models \varphi
    \\ \pv \models \varphi_1 \land \varphi_2 &  \Leftrightarrow & \pv \models
    \varphi_1 \textrm{ and } \pv \models \varphi_2
    \\ \pv \models \X \varphi & \Leftrightarrow & \pv^{(1)} \models \varphi
    \\ \pv \models \varphi_1 \U_T \varphi_2 & \Leftrightarrow & 
    \exists t \leq T. \ \big( \pv^{(t)} \models \varphi_2 \ \text{ and }  
    \\ & & \qquad (\forall t' < t. \ \pv^{(t')} \models \varphi_1 ) \big)
    \end{array} \vspace{-4pt}
\]
where $\pv^{(t)} (\cdot) = \pv (\cdot + t)$ is the $t$-time shift. 
Roughly, $\X \ap$ means $\ap$ holds \emph{next}, and $\ap_1 \U_T \ap_2$ means $\ap_1$ holds \emph{until} $\ap_2$ holds before $T$.
Other common logic operators are derived~by
\vspace{-4pt}
\[
\begin{split}
    & \textrm{Or: } \varphi \lor \varphi' \equiv \neg (\neg \varphi \land \neg \varphi'), \quad \textrm{Implies: } \varphi \Rightarrow \varphi' \equiv \neg \varphi \lor \varphi'
    \\ & \textrm{Finally: } \F_T \varphi \equiv \True \, \U_T \, \varphi, \qquad \textrm{Always: } \G_T \varphi \equiv \neg \F_T \neg \varphi.
\end{split} \vspace{-8pt}
\]
We also denote $\U_\infty$, $\F_\infty$, $\G_\infty$ by $\U$, $\F$, $\G$, respectively.

A large class of planning objectives can be expressed with LTL. 
For example, ``visiting $\ap_1$ and then $\ap_2$'' can be expressed by $\F (\ap_1 \land \F \ap_2)$; and ``always returning to the point $\ap_1$ within time $T$ after leaving it'' can be captured as $\G (\ap_1 \Rightarrow \F_T \ap_1)$.

\section{HyperLogics for Robotics Planing} \label{sec:motivation}

\begin{figure}[!t]
    \centering
    \begin{tikzpicture}[scale=2]
        \draw[->] (0, 0) to[out=-40] (2, 1) node[right] {$\pv_1$};
        \draw[fill] (0, 0) circle (0.03);
        \draw[->] (0, 0.1) to[out=0] (2, 0.6) node[right] {$\pv_2$};
        \draw[fill] (0, 0.1) circle (0.03);
        \node at (-0.2, 0.2) {Start};
        \draw[fill, opacity=0.5, dotted] (2, .8) ellipse (0.3 and 0.4) node[above, yshift=0.8cm] {Goal};
        \node at (0.3, 1.3) {Obstacle};
        \draw[pattern=north east lines] (0, 0.7) rectangle (0.8, 1.2);
        \draw[pattern=north east lines] (1.2, -0.2) rectangle (1.6, 0.5);
    
        \fill[magenta, opacity=0.2] (-.6, -.4) node[black, opacity=0.5, above right] {\small Region A} rectangle (1, .5);
        \fill[blue, opacity=0.2] plot coordinates {(-.6, .5) (-.6, 1.4) (1.4, 1.4) (1, .5)};
        \node[above right, black, opacity=0.5] at (-.6, .5) {\small Region B};

        \fill[green, opacity=0.2] plot coordinates {(1, .5) (1.4, 1.4) (2.5, 1.4) (2.5, -.4) (1, -.4)};
        \node[above left, black, opacity=0.5] at (2.4, -.4) {\small Region C};
    \end{tikzpicture}
     \vspace{-14pt}
    \caption{The use of hyperproperties in planning.}
    \vspace{-10pt}
    \label{fig:motivation}
\end{figure}
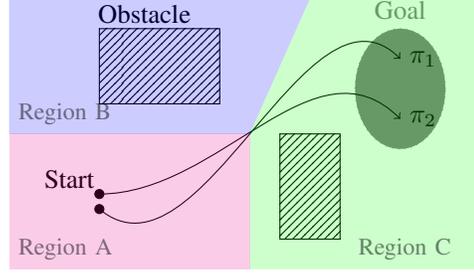

Although non-hyper temporal logics, such as LTL, are very expressive in temporal relations, they can only do so for individual paths; for example, whether a path $\pv_1$ in \cref{fig:motivation} reaches~the goal before hitting an obstacle.
However, many important planning objectives involve the interrelation between multiple paths, and thus cannot be expressed by these non-hyper temporal logics.
On the other hand, these objectives can be expressed in hyper-temporal logics, where explicit quantifications over different paths are allowed. In this section, we show the need for the use of hyperproperties in robotic planning on several motivating examples.

\paragraph*{Optimality of Synthesized Plans}
A well-known shortcoming of LTL-based planning is the lack of support for optimality.
For example, the objectives such as ``reaching the goal with the shortest time'' cannot be expressed in LTL, since formally defining such objectives implicitly involves comparison of the optimal path and other paths. 
On the other hand, with explicit path quantifications, the objective is achieved by finding a path $\pv$ such that
\vspace{-4pt}
\begin{equation} \label{eq:optimality_general}
    \begin{split}
        & \exists \pv. \Big( (\pv \textrm{ reaches goal}) \land 
        \\ & \quad \big( \forall \pv'. \big( (\pv' \textrm{ reaches goal}) \Rightarrow (\pv \textrm{ reaches goal}) \big) \Big).
    \end{split}    
\end{equation}

\vspace{-2pt}
\paragraph*{Robustness of Synthesized Plans}
A major concern~for open-loop planning is robustness of the derived strategy.
Specifically, (i) the assumed initial position may be inaccurate; or (ii) an action may not be executed correctly due to faults or attacks.
Yet, in many cases, the knowledge of the possible forms of inaccuracy, faults or attacks are available.
Hyper-temporal logics (e.g., HyperLTL) allow for incorporating this knowledge into the design objectives to preemptively synthesize strategies that are immune to those adversarial/environmental factors.
For example, a robust strategy under disturbance may be specified~by 
\begin{equation} \label{eq:robust_general}
    \begin{split}
        & \exists \pv \forall \pv'. \big( \pv' \textrm{ is derived by disturbing } \pv \big)
        \\ & \qquad \land \big(\pv \textrm{ and } \pv' \textrm{ reach goal} \big).
    \end{split}    
\end{equation}

\paragraph*{Planning with Privacy/Opacity} 
A problem that has recently attracted significant attentions
(e.g.,~\cite{gedik_LocationPrivacyMobile_2005,choudhary_DistributedTrajectoryEstimation_2016,li_CoordinatedMultirobotPlanning_2019}) 
is ensuring location privacy in mobile navigation -- i.e., keeping the individual locations private, even when they are partially shared to achieve coordinated planning (e.g.,~coverage).
Opacity ensures location privacy by requiring that for a planned path, there exists (at least) another different path, such that the shared partial location information is identical for the two paths; hence, they are anonymized.
An example is illustrated in \cref{fig:motivation}, where a path is partially observed by whether the robot is in Region A, B, or C or it reaches~the~goal.
Synthesizing an opaque planning strategy to reach the goal implies finding a path $\pv$ (or equivalently $\pv'$) such~that
\begin{equation} \label{eq:opacity_general}
\begin{split}
    & \exists \pv \exists \pv'. \big( \pv \textrm{ and } \pv' \textrm{ are different paths} \big) 
    \\ & \quad \land \big( \pv \textrm{ and } \pv' \textrm{ give identical observation} \big) 
    \\ & \quad \land \big( \pv \textrm{ and } \pv' \textrm{ reach goal} \big).
\end{split}
\end{equation}
The paths $\pv$ and $\pv'$ in \cref{fig:motivation} are examples of privacy-preserving paths, as they go through different regions, finally reaching the goal in the same pace; thus are~indistinguishable.

\section{Planning on Discrete Transition Systems} \label{sec:setup}

We consider the planning on a discrete domain, 
which can be either the full model of a complex workspace or its high-level abstraction derived from either simulation relation~\cite{dams_AbstractionAbstractionRefinement_2018} or random exploration~\cite{lavalle_PlanningAlgorithms_2006}.
On the domain, the robot motion is modeled by a DTS whose transitions are labeled by actions. 
 
\begin{definition} [DTS] \label{def:dts}
Given a set of atomic propositions $\aps$ a DTS is a tuple $\m = (\mss_\m, \mas_\m, \mts_\m, \ml_\m)$ where 
\begin{itemize}
    \item $\mss_\m$ is a set of \emph{states};
    \item $\mas_\m$ is a set of \emph{actions};
    \item $\mts_\m: \mss_\m \times \mas_\m \to \mss_\m$ is a partial \emph{transition function};
    \item $\ml_\m: \mss_\m \rightarrow 2^\aps$ is a labeling function determining the truth value of the atomic propositions on the states.
\end{itemize}
The subscript $\cdot_\m$ is omitted when it is clear from~the~context.
\end{definition}

A (open-loop) planning \textbf{strategy} $\str: \nat \rightarrow \mas$ on a DTS $\m$ is given by an infinite sequence of action;
clearly, for a finite time horizon planning problem, only a finite prefix of $\str$ takes effect.
Given an initial state $\msi \in \mss$ of the DTS, under the strategy $\str$, 
a \textbf{path} $\pv: \nat \rightarrow \mss$ can be generated, 
if $\pv(t+1) = \mts(\pv(t), \str(t))$ for all $t \in \nat$.
The planning task is then finding a path $\pv$ and corresponding strategy $\str$ such that objective $\varphi$ is satisfied.

\paragraph*{DTS Augmentation}
The DTS $\m$ from \cref{def:dts}~does not allow 
direct reference to the actions $\mas_\m$ using the atomic propositions, which are only associated to the states by $\ml_\m$.
To formalize our discussion, we introduce a mapping from $\m$ to an \emph{augmented DTS} $\s$ by encoding into states the actions taken previously.\footnote{Encoding the next action could incur unnecessary nondeterminism.}
The procedure is similar to the conversion from Moore machines into finite state automata.

\begin{definition} [Augmented DTS] \label{def:conversion}
 DTS $\s = (\mss_\s, \allowbreak \mas_\s, \mts_\s, \ml_\s)$ is an augmentation to a DTS $\m = (\mss_\m, \allowbreak \mas_\m, \mts_\m, \ml_\m)$,~if 
    \begin{itemize}
        \item $\mas_\s = \mas_\m$ and 
        $\{ \varepsilon \} \times \mss_\m 
        \subseteq \mss_\s 
        \subseteq (\mas_\m \cup \{ \varepsilon \}) \times \mss_\m$, where $\varepsilon$ is the empty sequence;         \item $(\ma_\m, \ms_\m) \in \mss_\s$ iff there exists $\ms_\m' \in \mss_\m$ such that $\mts_\m (\ms_\m', \ma_\m) = \ms_\m$;
        \item for any $\ma_\m \in \mas_\m$, $\big( (\cdot, \ms_\m'), \ma_\m, (\ma_\m, \ms_\m) \big) \in \mts_\s$ iff $\mts_\m (\ms_\m', \allowbreak \ma_\m) = \ms_\m$;
        \item for any $\ms_\s = (\ma_\m, \ms_\m) \in \mss_\s$, $\ml_\s (\ms_\s) = \ml_\m (\ms_\m) \cup \{\ma_\m, \ms_\m\}$;
    \end{itemize}
\end{definition}
For example, the DTS $\m$ in \cref{fig:initlalDTS} is augmented to the DTS $\s$ in \cref{fig:convertDTS}, where the actions L and R represent moving left and right.
Following \cref{def:conversion}, there is a correspondence between the paths of a DTS $\m$ and its augmented DTS $\s$, as formalized~below.
\begin{lemma} \label{lem:equivalence}
    Let $\pv_\m = \ms_\m(0) \ms_\m(1) \ldots$
    be a path of a DTS $\m$ under the strategy $\str = \ma_\m(0) \ma_\m(1) \ldots$. 
    Then $\pv_\s = (\varepsilon, \ms_\m(0)) (\ma_\m(0), \ms_\m(1)) \ldots \subseteq \mss_\s$ is a path of the equivalent augmented DTS $\s$ of $\m$; and \emph{vice versa}.  
\end{lemma}

From~\cref{def:conversion}, the states and actions of the initial DTS $\m$ are included in the labels of the augmented DTS $\s$.
Thus, planning on a DTS $\m$ with objectives specified over both its states and actions, can be mapped into planning on the augmented DTS $\s$ with objectives specified only over states, which can then be formally defined through the labels.

For quantifier-alternation-free hyper temporal objectives, a strategy can be synthesized by feeding the augmented DTS and the objective solely over states to an automata-based model checker (e.g., SPIN~\cite{holzmann_SpinModelChecker_2008}, PRISM~\cite{kwiatkowska_PRISMVerificationProbabilistic_2011}) with only a moderate modification (see~\cite{clarkson_TemporalLogicsHyperproperties_2014} for details).
For hyper objectives with alternating quantifiers, automata-based model checking is computational challenging. Hence, in this work, we use a symbolic model checking approach for the~synthesis.

\section{Planning from HyperLTL Specifications} \label{sec:logic}

To formally express hyperproperties-based planning objectives, in this section, we describe HyperLTL~\cite{clarkson_TemporalLogicsHyperproperties_2014}, which can be viewed as an extension of LTL to multiple paths.
We then show how HyperLTL can be used for planning with hyper-objectives, such as optimality, robustness and privacy.

\subsection{HyperLTL Syntax} \label{sub:hyperltl_syntax}
\vspace{-1pt}
HyperLTL enables reasoning about the interrelation~of~multiple paths by introducing a set of path variables $\Pi = \{\pv_1, \pv_2, \ldots\}$, where each path variable represents an individual path.
The atomic propositions are of the form $\ap^\pv$, where  the meaning of $\ap \in \aps$ is similar to LTL, and the superscript $\pv$ indicates that $\ap$ should be checked on the path~$\pv$.
These atomic propositions $\ap^\pv$ are concatenated by logic operators (e.g., $\neg$, $\land$, $\X$ and $\U_T$) as in LTL.
Finally, all the path variables in the objectives are quantified by $\exists$ or $\forall$. 
Formally, HyperLTL objectives are defined inductively by the syntax:
\vspace{-2pt}
\begin{align}
& \psi \Coloneqq \forall \pv. \ \psi
\ \vert \ \exists \pv. \ \psi
\ \vert \ \varphi \label{eq:hpltl_1}
\\ & \varphi \Coloneqq \ap^\pv
\ \vert \ \neg \varphi
\ \vert \ \varphi \land \varphi
\ \vert \ \X \varphi
\ \vert \ \varphi \U_T \varphi \label{eq:hpltl_2}
\end{align}
where 
$T \in \nat_\infty$,
$\ap \in \aps$, 
and $\pv \in \pvs$.
Other common logic operators are derived in the same way as in \cref{sub:ltl}.

\subsection{HyperLTL Semantics} \label{sub:hyperltl_semantics}
\vspace{-1pt}
As a HyperLTL objective may contain multiple path~variables, its satisfaction involves assigning concrete (infinite) paths to all these path variables. 
Therefore, we define $V: \pvs \rightarrow (2^\aps)^\omega$ as an \emph{assignment} for all possible path variables.~The satisfaction relation for the HyperLTL path formulas is then defined for $V$~by:
\begin{equation*} \begin{array}{l@{\hspace{1em}}c@{\hspace{1em}}l}
V \models \ap^\pv & \Leftrightarrow & \ap \in V(\pv) (0)
\\ V \models \neg \varphi & \Leftrightarrow & V \not\models \varphi
\\ V \models \varphi_1 \land \varphi_2 &  \Leftrightarrow & V \models
\varphi_1 \textrm{ and } V \models \varphi_2
\\ V \models \X \varphi & \Leftrightarrow & V^{(1)} \models \varphi
\\ V \models \varphi_1 \U_T \varphi_2 & \Leftrightarrow & 
\exists t \leq T. \ \big( V^{(t)} \models \varphi_2 \textrm{ and } 
\\ & & \quad (\forall t' < t. \ V^{(t')} \models \varphi_1) \big) 
\\ V \models \exists \pv. \ \psi & \Leftrightarrow & \textrm{there exists } \sig \in (2^\aps)^\omega, 
\\ & & \quad \textrm{such that } V[\pv \mapsto \sig] \models \psi
\\ V \models \forall \pv. \ \psi & \Leftrightarrow & \textrm{for all } \sig \in (2^\aps)^\omega, 
\\ & & \quad V[\pv \mapsto \sig] \models \psi \textrm{ holds}
\end{array}
\vspace{-2pt}
\end{equation*}
where $T \in \nat_\infty$ is a time horizon, and $V^{(t)}$ is the $t$-shift of the assignment $V$, defined by $\big(V^{(t)} (\pv)\big) \allowbreak = (V(\pv))^{(t)}$ for all path variables $\pv \in \pvs$.
Other logic operators, like $\lor$, $\Rightarrow$, $\G_T$, $\F_T$, $\U$, $\F$ and $\G$ are defined as for the LTL in \cref{sub:ltl}.

\begin{figure}[!t]
    \centering
    \begin{minipage}{0.15\textwidth}
        \centering
        \begin{tikzpicture}
            \node[draw, circle, inner sep=1] at (0, 0) (s1) {$\ms_1$};
            \node[draw, circle, inner sep=1] at (1, 0) (s2) {$\ms_2$};
            \node[draw, circle, inner sep=1] at (2, 0) (s3) {$\ms_3$};     
            
            \draw[->] (s1) to [bend left=15] node[above, inner sep=1] {{\scriptsize R}} (s2);
            \draw[->] (s2) to [bend left=15] node[above, inner sep=1] {{\scriptsize R}} (s3);

            \draw[->] (s2) to [bend left=15] node[below, inner sep=1] {{\scriptsize L}} (s1);
            \draw[->] (s3) to [bend left=15] node[below, inner sep=1] {{\scriptsize L}} (s2);
        \end{tikzpicture}
        \caption{Example DTS}
        \label{fig:initlalDTS}
    \vspace{5pt}
    \begin{tikzpicture}
            \node[left] at (0, 0) () {$\pv_1$};
            \draw[fill] (0, 0) circle (0.05) node[above] {$\ap_1$};
            \node at (0.5, 0) () {$...$};
            \draw (1.5, 0) circle (0.05);
            \draw[fill] (1, 0) circle (0.05) node[above] {$\ap_1$};
            \node at (2, 0) () {$...$};
            \draw[->] (0, 0) -- (.3, 0);
            \draw[->] (.7, 0) -- (.95, 0);
            \draw[->] (1, 0) -- (1.45, 0);
            \draw[->] (1.55, 0) -- (1.8, 0);
    
            \node[left] at (0, -.7) () {$\pv_2$};
            \draw (0, -.7) circle (0.05);
            \node at (0.5, -.7) () {$...$};
            \draw (1.5, -.7) circle (0.05);
            \draw[fill] (1, -.7) circle (0.05) node[above] {$\ap_2$};
            \node at (2, -.7) () {$...$};
            \draw[->] (0, -.7) -- (.3, -.7);
            \draw[->] (.7, -.7) -- (.95, -.7);
            \draw[->] (1, -.7) -- (1.45, -.7);
            \draw[->] (1.55, -.7) -- (1.8, -.7);
        \end{tikzpicture}
        \caption{Illustration for semantics of $\ap_1^{\pv_1} \U \ap_2^{\pv_2}$.}
        \label{fig:hyper until}
    \end{minipage}
    ~~
    \begin{minipage}{0.25\textwidth}
        \centering
        \begin{tikzpicture}
            \node[draw, ellipse, inner sep=0] at (0, 0) (s01) {$(\varepsilon, \ms_1)$};
            \node[draw, ellipse, inner sep=0] at (0, 1) (s02) {$(\varepsilon, \ms_2)$};
            \node[draw, ellipse, inner sep=0] at (0, 2) (s03) {$(\varepsilon, \ms_3)$};

            \node[draw, ellipse, inner sep=0] at (2, -0.5) (s1) {$(\mathrm{L}, \ms_1)$};
            \node[draw, ellipse, inner sep=0] at (2, 0.5) (s21) {$(\mathrm{R}, \ms_2)$};
            \node[draw, ellipse, inner sep=0] at (2, 1.5) (s22) {$(\mathrm{L}, \ms_2)$};
            \node[draw, ellipse, inner sep=0] at (2, 2.5) (s3) {$(\mathrm{R}, \ms_3)$};
            
            \draw[->] (s01) to node[above, at start, inner sep=1] {{\scriptsize R}} (s21);
            \draw[->] (s02) to node[above, at start, inner sep=1] {{\scriptsize R}} (s3);
            \draw[->] (s02) to node[below, at start, inner sep=1] {{\scriptsize L}} (s1);
            \draw[->] (s03) to node[below, at start, inner sep=1] {{\scriptsize L}} (s22);
            
            \draw[->] (s1.east) to [bend right=20] node[above, inner sep=1] {{\scriptsize R}} (s21.east);
            \draw[->] (s22.east) to [bend right=20] node[above, inner sep=1] {{\scriptsize R}} (s3.east);
            \draw[->] (s21.east) to [bend right=40] node[above, inner sep=1] {{\scriptsize R}} (s3.east);
            \draw[->] (s22.east) to [bend left=40] node[above, inner sep=1] {{\scriptsize L}} (s1.east);

            \draw[->] (s3.west) to [bend right=30] node[above, at start, inner sep=1] {{\scriptsize L}}  (s22.west);
            \draw[->] (s21.west)  to [bend right=20] node[above, at start, inner sep=1] {{\scriptsize L}} (s1.west);
        \end{tikzpicture}
        \vspace{-4pt}
        \caption{Augmented DTS}
        \label{fig:convertDTS}
    \end{minipage}
     \vspace{-15pt}
\end{figure}

\paragraph*{HyperLTL subsumes LTL} 
Any LTL objective can be expressed in HyperLTL.
For example, $(\s, \msi) \models \varphi$ for an~LTL objective $\varphi$ on a augmented DTS $\s$ with the initial state $\msi$, is expressed in HyperLTL by $V \models \varphi^\pv$, where $V(\pv)$ gives the path starting from $\msi$ of $\s$, and $\varphi^\pv$ means adding superscript $\pv$ to all atomic propositions in $\pv$.

\vspace{-1pt}
\paragraph*{HyperLTL is strictly more expressive than LTL}
Although the meaning of the logic operators in HyperLTL are similar to those in LTL, the ``until'' $\U$ (and the ``bounded until'' $\U_T$) in HyperLTL can be used between different paths.
For example, HyperLTL allows $\ap_1^{\pv_1} \U \ap_2^{\pv_2}$, meaning $\ap_1$ should hold on $\pv_1$ until $\ap_2$ should hold on $\pv_2$.
The satisfaction of $\ap_1^{\pv_1} \U \ap_2^{\pv_2}$ for the two paths $\pv_1$ and $\pv_2$ is illustrated in \cref{fig:hyper until}.

Also, HyperLTL allows alternating path quantifiers, like $\exists \pv_1 \forall \pv_2. \ \ap_1^{\pv_1} \U \ap_2^{\pv_2}$, which means that we look for a path $\pv_1$ such that for any path $\pv_2$ (possibly different from $\pv_1$), the objective $\ap_1^{\pv_1} \U \ap_2^{\pv_2}$ should be satisfied.
These ``until among multiple paths'' and ``exists such that for all'' cannot be expressed by LTL; thus, HyperLTL strictly subsumes~LTL.

\vspace{-4pt}
\subsection{HyperLTLf} \label{sub:hyperltlf}
\vspace{-4pt}
Most planning problems focus on finding a finite path,
while the semantics of HyperLTL
is defined for infinite paths.
We note that the semantics of HyperLTL 
can be adapted to finite paths to derive \emph{HyperLTLf}
in the same way as adapting LTL to LTLf~\cite{he_ReactiveSynthesisFinite_2017}.
Generally, HyperLTLf can be viewed as the finite-time fragment of HyperLTL.
Specifically, each finite path $\pv$ can be 
converted into an infinite path $\pv'$ by repeating the last entry.
We note that the semantics of HyperLTLf on the finite paths $\pv$
agrees with the semantics of HyperLTL on the infinite paths $\pv'$.
In~\cref{sec:method}, we introduce a symbolic synthesis method for handling the HyperLTLf objectives.

\vspace{-2pt}
\subsection{Applications of HyperLTL for Planning} \label{sec:app}
\vspace{-2pt}

We now show how HyperLTL (and HyperLTLf) can be used to formally express planning objectives~with~various types of robustness, optimality and privacy  properties~discussed in \cref{sec:motivation}.
As a running example, consider the inner navigation on a map of $3 \times 2$ rooms (\cref{fig:navigation}) where any two adjacent rooms are connected, $\ms_0$ is the \emph{start}, and the \emph{goal} is to reach either $\ms_4$ or $\ms_5$; 
a feasible path is shown with a thick solid line. 
The problem can be modeled by the DTS in \cref{fig:DTS}, where each state represents a room and the actions L, R, U and D denote moving left, right, up and~down.

\begin{figure}[!t]
\centering
\begin{minipage}{0.2\textwidth}
    \centering
    \begin{tikzpicture}
        \draw (0.,0.) -- (0.,2.);
        \draw (3.,2.) -- (0.,2.);
        \draw (3.,2.) -- (3.,0.);
        \draw (3.,0.) -- (0.,0.);
        \draw (1.,2.) -- (1.,1.6);
        \draw (1.,1.4) -- (1.,0.6);
        \draw (1.,0.4) -- (1.,0.);
        \draw (0.,1.) -- (0.4,1.);
        \draw (0.6,1.) -- (1.4,1.);
        \draw (1.6,1.) -- (2.4,1.);
        \draw (2.6,1.) -- (3.,1.);
        \draw (2.,2.) -- (2.,1.6);
        \draw (2.,1.4) -- (2.,0.6);
        \draw (2.,0.4) -- (2.,0.);

        \draw[thick] (0.5,0.5) -- (0.5,1.5);
        \draw[thick, -latex] (0.5,1.5) -- (1.4,1.5);

        \draw[dashed] (1.5,0.5) -- (1.5,1.5);
        \draw[dashed, -latex] (1.5,1.5) -- (2.5,1.5);
        
        \draw[thick, dotted] (0.5,0.5) -- (1.4,0.5);
        \draw[thick, dotted, -latex] (1.4,0.5) -- (1.4,1.5);        

        \node[below] at (0.2, 1) {$\ms_0$};
        \node[below] at (1.2, 1) {$\ms_1$};
        \node[below] at (2.2, 1) {$\ms_2$};
        \node[below] at (0.2, 2) {$\ms_3$};
        \node[below] at (1.2, 2) {$\ms_4$};
        \node[below] at (2.2, 2) {$\ms_5$};
    
        \node[above] at (0.5, 0) {start};
        \node[above] at (2, 1) {goal};

        \fill[red, opacity=0.5] (0, 0) rectangle (1, 1);
        \fill[green, opacity=0.5] (1, 1) rectangle (3, 2);
    \end{tikzpicture}
    \vspace{-8pt}
    \caption{Example workspace}
    \label{fig:navigation}
\end{minipage}
~~
\begin{minipage}{0.2\textwidth}
    \centering
    \begin{tikzpicture}
        \node[draw, circle, inner sep=1] at (0.5, 0.5) (s1) {$\ms_0$};
        \node[draw, circle, inner sep=1] at (1.5, 0.5) (s2) {$\ms_1$};
        \node[draw, circle, inner sep=1] at (2.5, 0.5) (s3) {$\ms_2$};
        \node[draw, circle, inner sep=1] at (0.5, 1.5) (s4) {$\ms_3$};
        \node[draw, circle, inner sep=1] at (1.5, 1.5) (s5) {$\ms_4$};
        \node[draw, circle, inner sep=1] at (2.5, 1.5) (s6) {$\ms_5$};        
        
        \draw[->] (s1) to [bend left=15] node[above, inner sep=1] {{\scriptsize R}} (s2);
        \draw[->] (s2) to [bend left=15] node[above, inner sep=1] {{\scriptsize R}} (s3);
        \draw[->] (s4) to [bend left=15] node[above, inner sep=1] {{\scriptsize R}} (s5);
        \draw[->] (s5) to [bend left=15] node[above, inner sep=1] {{\scriptsize R}} (s6);

        \draw[->] (s2) to [bend left=15] node[below, inner sep=1] {{\scriptsize L}} (s1);
        \draw[->] (s3) to [bend left=15] node[below, inner sep=1] {{\scriptsize L}} (s2);
        \draw[->] (s5) to [bend left=15] node[below, inner sep=1] {{\scriptsize L}} (s4);
        \draw[->] (s6) to [bend left=15] node[below, inner sep=1] {{\scriptsize L}} (s5);

        \draw[->] (s1) to [bend left=15] node[left, inner sep=1] {{\scriptsize U}} (s4);
        \draw[->] (s2) to [bend left=15] node[left, inner sep=1] {{\scriptsize U}} (s5);
        \draw[->] (s3) to [bend left=15] node[left, inner sep=1] {{\scriptsize U}} (s6);

        \draw[->] (s4) to [bend left=15] node[right, inner sep=1] {{\scriptsize D}} (s1);
        \draw[->] (s5) to [bend left=15] node[right, inner sep=1] {{\scriptsize D}} (s2);
        \draw[->] (s6) to [bend left=15] node[right, inner sep=1] {{\scriptsize D}} (s3);

    \end{tikzpicture}
    \caption{DTS model}
    \label{fig:DTS}
\end{minipage}
\vspace{-15pt}
\end{figure}

\subsubsection{Optimality}

LTL objectives cannot specify optimal strategies, such as shortest or longest paths,
as these implicitly involve comparison between multiple paths.
For example, a path $\pv_2$ reaches a goal set $g$ with the \emph{shortest time}, if it reaches $g$ before any other path $\pv_1$
HyperLTL can specify this property by $\F (g^{\pv_2} \Rightarrow \F g^{\pv_1})$.
Thus, the objective of synthesizing a strategy for reaching $g$ from an initial state $\msi$ \textbf{the shortest path} is given by
\begin{equation} \label{eq:shortest}
    \exists \pv_2 \forall \pv_1. \ \big( \msi^{\pv_1} \land \msi^{\pv_2} \big) \land (\F g^{\pv_2}) 
    \land \big( \F ( g^{\pv_2} \Rightarrow \F g^{\pv_1} ) \big);
\end{equation}
and the objective for \textbf{the longest path} is captured by
\begin{equation} \label{eq:longest}
    \exists \pv_2 \forall \pv_1. \ \big( \msi^{\pv_1} \land \msi^{\pv_2} \big) 
    \land \big( \F ( g^{\pv_1} \Rightarrow \F g^{\pv_2} ) \big).
\end{equation}
Note that assuming the paths are no longer than some $T \in \nat_\infty$, specifications~\eqref{eq:shortest} and~\eqref{eq:longest} can be 
equivalently expressed by replacing the ``unbounded'' $\F$ with the ``bounded''~$\F_T$. 
In \cref{fig:navigation}, the path $\pv_2$ in solid line satisfies the HyperLTL objective~\eqref{eq:shortest} hence, is the shortest path.

\vspace{4pt}
\subsubsection{Robustness}

HyperLTL enables capturing requirements for synthesizing a planning strategy that, with its initial objectives achieved, is also robust to various types of uncertainties, and even faults and adversarial factors. 
Generally, let $\varphi$ be an LTL objective to be \textbf{\textit{robustly}} satisfied, and $\mathrm{cls}_\msi (\pv_1, \pv_2)$ and $\mathrm{cls}_\mas (\pv_1, \pv_2)$ be notions of ``closeness'' of the initial states and actions.
Robust planning is defined by an objective that there exists a path $\pv_1$ such that, for any other path $\pv_2$ close to $\pv_1$, the objective $\varphi$ should still be~satisfied:
\begin{equation} \label{eq:robustness}
    \exists \pv_1 \forall \pv_2. \ \left(\mathrm{cls}_\msi (\pv_1, \pv_2) \land \mathrm{cls}_\mas (\pv_1, \pv_2)\right) \Rightarrow \big( \varphi^{\pv_1} \land \varphi^{\pv_2} \big),
\end{equation}
where $\varphi^\pv$ is derived by replacing all the atomic propositions $\ap$ in $\varphi$ by $\ap^\pv$.
Depending on the different sources of uncertainty, we highlight the following notions of robustness.

\vspace{2pt}
\noindent\textbullet~\textbf{Initial-state robustness} when the uncertainty comes not fully knowing the initial state -- i.e, from replacing a predefined initial state $\msi$ to an arbitrary state from a set $S_0$. In this case, we capture the objective of synthesizing an initial-state robust strategy  for a time horizon $T \in \nat_\infty$ for an LTL objective $\varphi$, as the HyperLTL formula   
     \begin{equation} \label{eq:is robustness}
    \begin{split}
        \exists \pv_1 \forall \pv_2. \big( \msi^{\pv_1} \hspace{-2pt} \land \hspace{-2pt} S_0^{\pv_2} \big) \hspace{-2pt}\land \hspace{-2pt} ( \varphi^{\pv_1} \hspace{-2pt} \land \hspace{-2pt} \varphi^{\pv_2} )
         \land \big( \G_T (\ma^{\pv_1} = \ma^{\pv_2}) \big) 
    \end{split}
\end{equation}
Note that in~\eqref{eq:is robustness} and the formulas below, ``$=$'' is not an arithmetic relation, but a notation simplification: $\ma^{\pv_1} = \ma^{\pv_2}$ stands for $\bigwedge_{\ma \in \mas} (\ma^{\pv_1} \land \ma^{\pv_2})$.

\noindent\textbullet~\textbf{Action robustness} when the uncertainty comes from~control faults -- e.g., from replacing at most one action with another arbitrary action. Then, in HyperLTL, we capture the objective of synthesizing an action robust strategy for a time horizon $T \in \nat_\infty$ for an LTL objective $\varphi$ as the objective
    \begin{equation} \label{eq:a robustness}
    \begin{split} 
        & \exists \pv_1 \forall \pv_2. \ \big( \msi^{\pv_1} \land \msi^{\pv_2} \big) \land ( \varphi^{\pv_1} \land \varphi^{\pv_2} )
        \\ & \qquad \land \G_T \big( (\ma^{\pv_1} \neq \ma^{\pv_2}) \Rightarrow \big( \X \G_T (\ma^{\pv_1} = \ma^{\pv_2}) \big) \big)  
    \end{split}
    \end{equation}
Robustness to other types of action uncertainties (disturbances), such as at most $N$ or no $N$ successive replacements, can be similarly expressed in HyperLTL.

In \cref{fig:navigation}, the strategy (first up then right) shown by the solid line is initial-state robust~if the initial state is uncertain between $\ms_0$ and $\ms_1$, because the same strategy generates another feasible path shown in the dashed line.
However, the strategy is not action robust as the robot will not reach the goal, if either the first or second action is~replaced.

\vspace{4pt}
\subsubsection{Privacy/Opacity}

Let $\mathrm{sec} (\cdot)$ be a \emph{secret} and $\mathrm{obs} (\cdot)$ be a (partial) \emph{observation} on a path.
A opaque strategy satisfies that there exist at least two paths with the same observation but bearing different secrets, such that the secret of each path cannot be identified exactly only from the observation -- i.e., 
\begin{equation} \label{eq:opacity}
    \exists \pv_1 \exists \pv_2. \big( \mathrm{sec}(\pv_1) \neq \mathrm{sec}(\pv_2) \big) \land \big( \mathrm{obs}(\pv_1) = \mathrm{obs}(\pv_2) \big).
\end{equation} 
Depending on the specific forms of the secrets and observations, we consider the following notions of privacy/opacity.

\vspace{2pt}
\noindent\textbullet~\textbf{Initial-state opacity for fixed strategy}~\cite{saboori_VerificationInitialstateOpacity_2013}: 
    Let the secret be the initial state of the path and observe whether the robot finally reaches a goal set $g$. 
    Then, the objective of synthesizing an initial-state opaque strategy from the initial state $\msi$ for a time horizon $T \in \nat_\infty$ can be captured by     \begin{equation} \label{eq:is opacity}
    \begin{split}
        & \exists \pv_1 \exists \pv_2. \big( \msi^{\pv_1} \land (\neg \msi^{\pv_2}) \big) 
        \\ & \qquad \land \big( \G_T (\ma^{\pv_1} = \ma^{\pv_2}) \big) 
        \land \big( (\F_T g^{\pv_1}) \land (\F_T g^{\pv_2}) \big).
    \end{split}
    \end{equation}

\vspace{2pt}
\noindent\textbullet~\textbf{Current-state opacity}~\cite{yin_NewApproachSynthesizing_2015}: 
    Let the secret be the synthesized strategy, and the observation be the initial state and whether the path is \emph{currently} in a set $o$. Then, the HyperLTL objective of synthesizing a current-state opaque strategy from the initial state $\msi$ for a time horizon $T \in \nat_\infty$ is captured~as
    \begin{equation} \label{eq:cs opacity}
    \begin{split}
        & \exists \pv_1 \exists \pv_2. \big( \msi^{\pv_1} \land \msi^{\pv_2} \big) \land 
        \\ & \qquad \land \big( \neg \G_T (\ma^{\pv_1} = \ma^{\pv_2}) \big) 
        \land \big( \G_T (o^{\pv_1} = o^{\pv_2}) \big).
    \end{split}
    \end{equation}
    The above formula requires that there are two different paths $\pv_1$ and $\pv_2$, generated by two different strategies, such that they give the same observations; and any one of these two is a current-state opaque strategy.

In \cref{fig:navigation}, it is easy to check that the strategy shown~by the solid line is not initial-state opaque.
However, the strategy is current state private because of the existence of the strategy shown by the dotted line.

\section{Strategy Synthesis} \label{sec:method}

In this section, we introduce a symbolic approach for synthesizing strategies 
for HyperLTL objectives.
Similarly to existing work on symbolic planning, such as~\cite{biere_BoundedModelChecking_2003,he_EfficientSymbolicReactive_2019,he_ReactiveSynthesisFinite_2017} and references therein,
we consider HyperLTL objectives with bounded time horizons
(effectively, HyperLTLf from \cref{sub:hyperltlf})
via the following three steps.
First, we identify a required time horizon for synthesizing a strategy for a considered objective, as introduced in \cref{sub:time horizon}.
Then, as presented in \cref{sub:model conversion for SMT}, by replacing the $\exists$ and $\forall$ quantifications over paths to that over a finite sequence of states and actions within the required horizon time, we convert the HyperLTL objective and the constraint of the DTS model on its path, into first-order logic formulas.
Finally, the conjunction of the two formulas representing the system model and the synthesis objective, is solved using an off-the-shelf SMT solver.

\subsection{Computing Required Time Horizon} \label{sub:time horizon}

A HyperLTL objective contains multiple paths. Therefore, unlike with LTL formulas, the required time horizon may be different among the utilized path variables.
Specifically, let $H(\varphi, \pv)$ be the required time horizon for a path variable $\pv$ in a HyperLTL objective~$\varphi$.
For an atomic proposition $\ap^{\pv}$, the required time horizon is $0$, if the path variable $\pv$ appears in it, and $-\infty$ otherwise -- i.e., formally we define
\vspace{-2pt}
\[
    H(\ap^{\pv}, \pv') = \begin{cases}
        0 & \textrm{ if } \pv' = \pv \\
        -\infty & \textrm{ otherwise.} \\
    \end{cases}.
\]
Furthermore, every ``next'' and ``until'' temporal operator employed in formula $\varphi$ changes the required time horizon as captured by the following recursive rules 
\vspace{-2pt}
\[
    \begin{split}
        & H(\X \varphi, \pv) = H(\varphi, \pv) + 1,
        \\ & H(\varphi_1 \U_T \varphi_2, \pv) = \max\{ H(\varphi_1, \pv), H(\varphi_2, \pv) \} + T.
    \end{split}  
\]
Finally, the negation and quantification for path variables do not change the required time horizon -- i.e., for any $\pv$, \[
\begin{split}
& H(\neg \varphi, \pv) = H(\varphi, \pv), 
\\ & H(\varphi_1 \land \varphi_2, \pv) = \max\{ H(\varphi_1, \pv), H(\varphi_2, \pv) \},
\\ & H(\exists \pv'. \ \psi, \pv) = H(\psi, \pv), \quad H(\forall \pv'. \ \psi, \pv) = H(\psi, \pv).
\end{split}
\]
In the above rules, we follow the convention that $x + (- \infty) = - \infty$ and $\max\{x, - \infty\} = x$ for any $x \in \nat$.
Also, for HyperLTLf objective $\varphi$, if a path variable $\pv$ appears in $\varphi$, then its time horizon $H(\varphi, \pv)$ is finite; otherwise, $H(\varphi, \pv) = -\infty$.
Finally, as an example, $H(\exists \pv_2 \forall \pv_1. \ \big( \msi^{\pv_1} \land \msi^{\pv_2} \big) \land \big( \F_T ( g^{\pv_2} \Rightarrow \F_T g^{\pv_1} ) \big), \pv_1) 
= 
H( \F_T ( g^{\pv_2} \Rightarrow \F_T g^{\pv_1} ), \pv_1)
= 
H( \F_T g^{\pv_1} , \pv_1) + T 
= 2 T$.

\subsection{Model Conversion for SMT-based Synthesis} \label{sub:model conversion for SMT}

Consider a HyperLTLf objective of the general form $\mathsf{Q}_1 \pv_1 \ldots \mathsf{Q}_n \pv_n. \ \varphi$, where $\mathsf{Q}_i \in \{\exists, \forall\}$ for $i \in [n]$.
For each path variable $\pv_i$, let its required time horizon in the objective be $H_i = H(\varphi, \pv_i)$, where $H(\cdot, \cdot)$ is computed as described in \cref{sub:time horizon}.
Then, the path quantifications over $\pv_i$ is equivalently represented by its initial state $\ms_i(0)$ and its actions $\ma_i(0) \ldots \ma_i(H_i-1)$.
Since the path should be generated from a DTS $\m$ as introduced in \cref{def:dts}, it should satisfy that
\begin{equation} \label{eq:pi}
    P_i = \bigwedge\nolimits_{t \in [H_i]} \big( \ms_i(t) = \mts_\m(\ms_i(t-1), \ma_i(t-1)) \big), \quad i \in [n],
\end{equation}
where $\ms_i(0) \in \mss_\m$ and $\ma_i(0), \ldots, \ma_i(H_i - 1) \in \mas_\m$ are viewed as variables.
Equivalently, this constraint~\eqref{eq:pi} can be generated from the augmented DTS $\s$.
Finally, for each $i \in [n]$, the path quantification $\mathsf{Q}_i \pv_i$ is equivalently represented~by
\begin{equation} \label{eq:qi}
    [\mathsf{Q}_i \pv_i] = \mathsf{Q}_i \ms_i(0) \mathsf{Q}_i \ma_i(0) \ldots \mathsf{Q}_i \ma_i(H_1 - 1). 
\end{equation}
Consequently, the planning strategy should satisfy the following first-order logic formula
\begin{equation} \label{eq:goal}
    [\mathsf{Q}_1 \pv_1] \ldots [\mathsf{Q}_n \pv_n]. \big( \bigwedge\nolimits_{i \in [n]} P_i \big) \land \varphi,
\end{equation}
where $P_i$ is introduced in~\eqref{eq:pi} and $[\mathsf{Q}_i \pv_i]$ for $i \in [n]$ is defined in~\eqref{eq:qi}. 
This formula~\eqref{eq:goal} can be directly solved by~SMT~solvers, such as~Z3, Yices~\cite{dutertre_YICESSMTSolver_2006} or CVC4~\cite{barrett_CVC4_2011}.

\section{Implementation and Case Studies} \label{sec:cases}

We implemented the described symbolic synthesis method. Specifically, the conversion from the DTS and HyperLTL objectives to first-order logic expressions is implemented in Python.
Then, the expressions of the form~\eqref{eq:goal} are solved by Z3 SMT solver~\cite{demoura_Z3EfficientSMT_2008}.
The source code is available at~\cite{mphyper_HttpsGitlabOit_2019}.

\begin{figure}[!t]
    \centering
    \begin{minipage}{.22\textwidth}
        \centering
        \includegraphics[width=0.8\linewidth,]{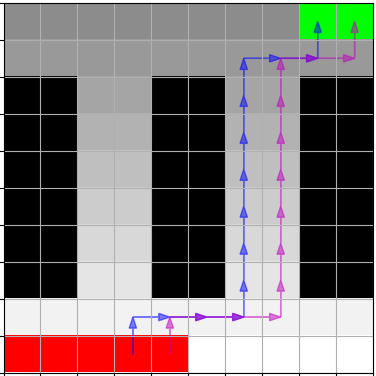}
        \vspace{-10pt}
        \caption{Strategy for initial-state opacity \eqref{eq:is opacity} for time horizon $T=20$; the synthesis time is $<0.14$s.}
        \label{fig:Initial-state opacity}
    \end{minipage}    \hspace{18pt}
    \begin{minipage}{.22\textwidth}
        \centering
        \includegraphics[width=0.8\linewidth]{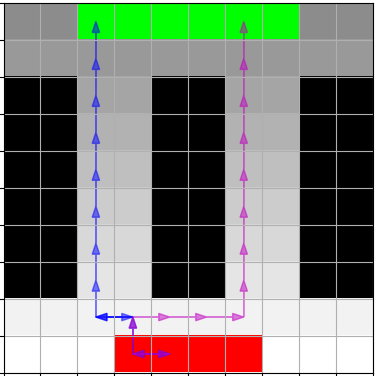}
        \vspace{-10pt}
        \caption{Strategy for current-state opacity \eqref{eq:cs opacity} for time horizon $T=20$; the synthesis time is $<0.12$s.}
        \label{fig:Current-state opacity}
    \end{minipage}
\end{figure}

\begin{figure}[!t]
    \centering
    \begin{minipage}{.22\textwidth}
        \centering
        \includegraphics[width=0.8\linewidth]{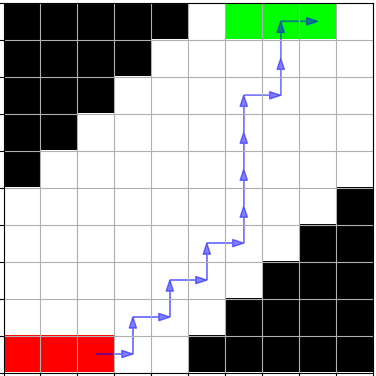}
         \vspace{-10pt}
        \caption{Strategy for initial-state~robustness \eqref{eq:is robustness} for time horizon $T=20$; the synthesis time is $<0.24$s.}
        \label{fig:Initial-state robustness}
    \end{minipage}     \hspace{18pt}
    \begin{minipage}{.22\textwidth}
        \centering
        \includegraphics[width=0.8\linewidth]{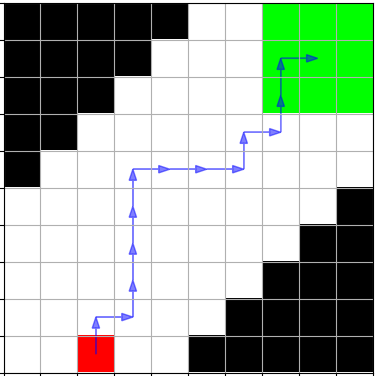}
       \vspace{-10pt}
        \caption{Strategy for action robustness \eqref{eq:a robustness} for time horizon $T=20$; the synthesis time is $<0.15$s.}
        \label{fig:Action robustness}
    \end{minipage}
\end{figure}

\begin{figure}[!t]
    \centering
    \includegraphics[width=0.4\linewidth]{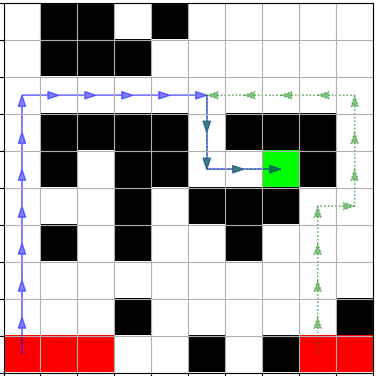}
    \vspace{-8pt}
    \caption{Synthesized strategy (in blue) for shortest path \eqref{eq:shortest} for time horizon $T=20$; the synthesis time is $<0.15$s.}
    \label{fig:Shortest path}
\end{figure}

The implemented method is evaluated on several planning problems of a mobile robot on grid worlds with obstacles, as illustrated in Figures~\ref{fig:Initial-state opacity}-\ref{fig:Shortest path}, where the black, white, red, and green colors stand for obstacles, allowable states, start states and goal states, respectively.
At each step, the robot can move up, down, left or right; upon hitting an obstacle, the objective immediately fails.
We focused on the HyperLTL objectives discussed in \cref{sec:app} for a finite horizon $T \in \nat$.

The feasible paths on $10 \times 10$ grids for several objectives are illustrated in Figures~\ref{fig:Initial-state opacity} to~\ref{fig:Shortest path}.
For the \textbf{privacy/opacity objectives} from~\eqref{eq:is opacity} and~\eqref{eq:cs opacity}, let the partial observation be the row number of the current state of the robot, depicted by the color gradient from bottom to top.
In \cref{fig:Initial-state opacity}, the synthesized blue path is initial-state opaque, as there exists the red path that follows the same strategy, yields the same observation along the path, and reaches the goal, but starts from a different state.
In \cref{fig:Current-state opacity}, the synthesized blue path is current-state opaque, as there exists the red path that starts from the same state, yields the same observation along the path, and reaches the goal, but follows a different path.
It is worth noting that to achieve current-state opacity, the synthesized blue path actually waits at the bottom row for two steps to ensure another indistinguishable path can catch~up.

For the \textbf{robustness objectives} from~\eqref{eq:is robustness} and~\eqref{eq:a robustness}, the synthesized blue path in \cref{fig:Initial-state robustness} is initial-state robust, meaning the corresponding strategy is feasible for any initial state in the red region.
The synthesized blue path in \cref{fig:Action robustness} is action robust, as defined in~\eqref{eq:a robustness} -- i.e., the corresponding strategy is feasible for any single  action replacement. 
In both cases, the blue path avoids getting close to the obstacles, in case for initial state inaccuracy or action errors.
Finally, for the \textbf{optimality objective} from~\eqref{eq:shortest}, the synthesized blue path in \cref{fig:Shortest path} shows the shortest path from a red state to the green state (another shortest path is the dotted green path).

\cref{tb:results} presents the times for synthesizing strategies for the above cases, as well as problems on larger grids; all computations were done on an Intel i7-7820X CPU~@3.60GHz and RAM 32GB (only one core was used).
The time horizon $T$ was chosen such that the goal is reachable to prevent~easy fails.
As shown in the table, the strategy synthesis of the HyperLTL objectives can be performed in a reasonable amount of time even on nontrivial problems.
As expected, there is an increase in synthesis times as the grid size and time horizon increase, since in the worst case, the size of the first-order logic formula~\eqref{eq:goal} to be evaluated (i.e., solved) by Z3, can grow exponentially with the time horizon and the number of states.

\begin{table}[!t]
    \centering
    \caption{Symbolic synthesis times for HyperLTL objectives on grid worlds with obstacles. ISO, CSO, ISR, AR and S stands for initial-state opacity, current-state opacity, initial-state robustness, action robustness and shortest path, respectively.}
    \label{tb:results}
    \begin{tabular}{cccc|cccc}
        \hline  \hline
        Grid   & Obj. & $T$ & Time (s) & Grid   & Obj. & $T$ & Time (s) \\
        \hline
        $10^2$ & ISO       & $20$ & $0.14$ & $20^2$ & ISO       & $40$ & $5.2$ \\
        $10^2$ & CSO       & $20$ & $0.12$ & $20^2$ & CSO       & $40$ & $2.9$ \\
    	$10^2$ & ISR       & $20$ & $0.24$ & $20^2$ & ISR       & $40$ & $4.7$ \\
    	$10^2$ & AR        & $20$ & $0.15$ & $20^2$ & AR        & $40$ & $5.5$ \\
        $10^2$ & SP        & $20$ & $0.15$ & $20^2$ & SP        & $40$ & $5.0$ \\

        $40^2$ & ISO       & $80$ & $30$  & $60^2$  &    ISO   &  $120$   &   $382$ \\
        $40^2$ & CSO       & $80$ & $24$  & $60^2$  &    CSO   &  $120$   &   $191$ \\
    	$40^2$ & ISR       & $80$ & $49$  & $60^2$  &    ISR   &  $120$   &   $320$ \\
    	$40^2$ & AR        & $80$ & $38$  & $60^2$  &    AR    &  $120$   &   $306$ \\
        $40^2$ & SP        & $80$ & $172$ & $60^2$  &    SP    &  $120$   &   $244$ \\
        \hline  \hline
    \end{tabular}
    \vspace{-15pt}
\end{table}
 
\section{Conclusion} \label{sec:conc}

In this work, we proposed the use of HyperLTL in planning for specifying objectives involving the interrelation of multiple paths, such as optimality, robustness and privacy/opacity, which cannot be expressed by the widely used temporal logics, such as linear temporal logic (LTL).  
We showed how those hyperproperties can be expressed by HyperLTL, which is an extension of LTL to multiple paths.
Then, we  introduced a method for symbolic synthesis of high-level planning strategies from such HyperLTL objectives, using off-the-shelf tools. 
Finally, we evaluated the proposed method on several planning case studies.

\bibliographystyle{IEEEtran}

\begin{thebibliography}{10}
\providecommand{\url}[1]{#1}
\csname url@samestyle\endcsname
\providecommand{\newblock}{\relax}
\providecommand{\bibinfo}[2]{#2}
\providecommand{\BIBentrySTDinterwordspacing}{\spaceskip=0pt\relax}
\providecommand{\BIBentryALTinterwordstretchfactor}{4}
\providecommand{\BIBentryALTinterwordspacing}{\spaceskip=\fontdimen2\font plus
\BIBentryALTinterwordstretchfactor\fontdimen3\font minus
  \fontdimen4\font\relax}
\providecommand{\BIBforeignlanguage}[2]{{\expandafter\ifx\csname l@#1\endcsname\relax
\typeout{** WARNING: IEEEtran.bst: No hyphenation pattern has been}\typeout{** loaded for the language `#1'. Using the pattern for}\typeout{** the default language instead.}\else
\language=\csname l@#1\endcsname
\fi
#2}}
\providecommand{\BIBdecl}{\relax}
\BIBdecl

\bibitem{fainekos_TemporalLogicMotion_2005}
G.~Fainekos, H.~{Kress-Gazit}, and G.~Pappas,
  ``\BIBforeignlanguage{en}{Temporal {{Logic Motion Planning}} for {{Mobile
  Robots}}},'' in \emph{\BIBforeignlanguage{en}{Proceedings of the 2005 {{IEEE
  International Conference}} on {{Robotics}} and {{Automation}}}}.\hskip 1em
  plus 0.5em minus 0.4em\relax {IEEE}, 2005, pp. 2020--2025.

\bibitem{plaku_MotionPlanningTemporallogic_2016}
E.~Plaku and S.~Karaman, ``\BIBforeignlanguage{en}{Motion planning with
  temporal-logic specifications: {{Progress}} and challenges},''
  \emph{\BIBforeignlanguage{en}{AI Communications}}, vol.~29, no.~1, pp.
  151--162, 2016.

\bibitem{elfar_cav19}
M.~Elfar, Y.~Wang, and M.~Pajic, ``Security-aware synthesis using
  delayed-action games,'' in \emph{Computer Aided Verification (CAV)}.\hskip
  1em plus 0.5em minus 0.4em\relax Springer International Publishing, 2019, pp.
  180--199.

\bibitem{kress-gazit_TemporalLogicBasedReactiveMission_2009}
H.~{Kress-Gazit}, G.~E. Fainekos, and G.~J. Pappas,
  ``Temporal-{{Logic}}-{{Based Reactive Mission}} and {{Motion Planning}},''
  \emph{IEEE Transactions on Robotics}, vol.~25, no.~6, pp. 1370--1381, 2009.

\bibitem{moarref_ReactiveSynthesisRobotic_2018}
S.~Moarref and H.~{Kress-Gazit}, ``\BIBforeignlanguage{en}{Reactive
  {{Synthesis}} for {{Robotic Swarms}}},'' in
  \emph{\BIBforeignlanguage{en}{Formal {{Modeling}} and {{Analysis}} of {{Timed
  Systems}}}}.\hskip 1em plus 0.5em minus 0.4em\relax {Springer International
  Publishing}, 2018, vol. 11022, pp. 71--87.

\bibitem{kress-gazit_WhereWaldoSensorBased_2007}
H.~{Kress-Gazit}, G.~E. Fainekos, and G.~J. Pappas, ``Where's {{Waldo}}?
  {{Sensor}}-{{Based Temporal Logic Motion Planning}},'' in \emph{Proceedings
  2007 {{IEEE International Conference}} on {{Robotics}} and {{Automation}}},
  2007, pp. 3116--3121.

\bibitem{he_ReactiveSynthesisFinite_2017}
K.~He, M.~Lahijanian, L.~E. Kavraki, and M.~Y. Vardi,
  ``\BIBforeignlanguage{en}{Reactive synthesis for finite tasks under resource
  constraints},'' in \emph{\BIBforeignlanguage{en}{2017 {{IEEE}}/{{RSJ
  International Conference}} on {{Intelligent Robots}} and {{Systems}}
  ({{IROS}})}}.\hskip 1em plus 0.5em minus 0.4em\relax {IEEE}, 2017, pp.
  5326--5332.

\bibitem{kantaros_GlobalPlanningMultiRobot_2016}
Y.~Kantaros and M.~M. Zavlanos, ``Global {{Planning}} for {{Multi}}-{{Robot
  Communication Networks}} in {{Complex Environments}},'' \emph{IEEE
  Transactions on Robotics}, vol.~32, no.~5, pp. 1045--1061, 2016.

\bibitem{kantaros_TemporalLogicTask_2019}
Y.~Kantaros, M.~Guo, and M.~M. Zavlanos, ``Temporal {{Logic Task Planning}} and
  {{Intermittent Connectivity Control}} of {{Mobile Robot Networks}},''
  \emph{IEEE Transactions on Automatic Control}, pp. 1--1, 2019.

\bibitem{kantaros_STyLuSTemporalLogic_2018}
Y.~Kantaros and M.~M. Zavlanos, ``{{STyLuS}}: {{A Temporal Logic Optimal
  Control Synthesis Algorithm}} for {{Large}}-{{Scale Multi}}-{{Robot
  Systems}},'' \emph{arXiv:1809.08345 [cs]}, 2018.

\bibitem{farahani_RobustModelPredictive_2015}
S.~S. Farahani, V.~Raman, and R.~M. Murray, ``\BIBforeignlanguage{en}{Robust
  {{Model Predictive Control}} for {{Signal Temporal Logic Synthesis}}},''
  \emph{\BIBforeignlanguage{en}{IFAC-PapersOnLine}}, vol.~48, no.~27, pp.
  323--328, Jan. 2015.

\bibitem{lindemann_RobustMotionPlanning_2017}
L.~Lindemann and D.~V. Dimarogonas, ``Robust motion planning employing signal
  temporal logic,'' in \emph{2017 {{American Control Conference}} ({{ACC}})},
  May 2017, pp. 2950--2955.

\bibitem{wolff_OptimalControlNondeterministic_2013}
E.~M. Wolff, U.~Topcu, and R.~M. Murray, ``\BIBforeignlanguage{en}{Optimal
  control of non-deterministic systems for a computationally efficient fragment
  of temporal logic},'' in \emph{\BIBforeignlanguage{en}{52nd {{IEEE
  Conference}} on {{Decision}} and {{Control}}}}.\hskip 1em plus 0.5em minus
  0.4em\relax {Firenze}: {IEEE}, Dec. 2013, pp. 3197--3204.

\bibitem{jing_ShortcutEvilDoor_2013}
G.~Jing, R.~Ehlers, and H.~{Kress-Gazit}, ``Shortcut through an evil door:
  {{Optimality}} of correct-by-construction controllers in adversarial
  environments,'' in \emph{2013 {{IEEE}}/{{RSJ International Conference}} on
  {{Intelligent Robots}} and {{Systems}}}, Nov. 2013, pp. 4796--4802.

\bibitem{clarkson_Hyperproperties_2008}
M.~R. Clarkson and F.~B. Schneider,
  ``\BIBforeignlanguage{en}{Hyperproperties},'' in
  \emph{\BIBforeignlanguage{en}{2008 21st {{IEEE Computer Security Foundations
  Symposium}}}}.\hskip 1em plus 0.5em minus 0.4em\relax {IEEE}, 2008, pp.
  51--65.

\bibitem{clarkson_TemporalLogicsHyperproperties_2014}
M.~R. Clarkson, B.~Finkbeiner, M.~Koleini, K.~K. Micinski, M.~N. Rabe, and
  C.~S{\'a}nchez, ``\BIBforeignlanguage{en}{Temporal {{Logics}} for
  {{Hyperproperties}}},'' in \emph{\BIBforeignlanguage{en}{Principles of
  {{Security}} and {{Trust}}}}.\hskip 1em plus 0.5em minus 0.4em\relax
  {Springer Berlin Heidelberg}, 2014, vol. 8414, pp. 265--284.

\bibitem{lavalle_PlanningAlgorithms_2006}
S.~M. LaValle, \emph{\BIBforeignlanguage{en}{Planning {{Algorithms}}}}.\hskip
  1em plus 0.5em minus 0.4em\relax {Cambridge University Press}, 2006.

\bibitem{lu_FrameworkModelChecking_2014}
Y.~Lu, Y.~Guan, X.~Li, R.~Wang, and J.~Zhang, ``A framework of model checking
  guided test vector generation for the {{6DOF}} manipulator,'' in \emph{2014
  {{IEEE International Conference}} on {{Robotics}} and {{Automation}}
  ({{ICRA}})}, 2014, pp. 4262--4267.

\bibitem{schillinger_MultiobjectiveSearchOptimal_2017}
P.~Schillinger, M.~B{\"u}rger, and D.~V. Dimarogonas, ``Multi-objective search
  for optimal multi-robot planning with finite {{LTL}} specifications and
  resource constraints,'' in \emph{2017 {{IEEE International Conference}} on
  {{Robotics}} and {{Automation}} ({{ICRA}})}, 2017, pp. 768--774.

\bibitem{kupferman_AutomataTheoryModel_2018}
O.~Kupferman, ``\BIBforeignlanguage{en}{Automata {{Theory}} and {{Model
  Checking}}},'' in \emph{\BIBforeignlanguage{en}{Handbook of {{Model
  Checking}}}}.\hskip 1em plus 0.5em minus 0.4em\relax {Springer International
  Publishing}, 2018, pp. 107--151.

\bibitem{finkbeiner_AlgorithmsModelChecking_2015}
B.~Finkbeiner, M.~N. Rabe, and C.~S{\'a}nchez,
  ``\BIBforeignlanguage{en}{Algorithms for {{Model Checking HyperLTL}} and
  {{HyperCTL}}*},'' in \emph{\BIBforeignlanguage{en}{Computer {{Aided
  Verification}}}}.\hskip 1em plus 0.5em minus 0.4em\relax {Springer
  International Publishing}, 2015, pp. 30--48.

\bibitem{biere_SATBasedModelChecking_2018}
A.~Biere and D.~Kr{\"o}ning, ``\BIBforeignlanguage{en}{{{SAT}}-{{Based Model
  Checking}}},'' in \emph{\BIBforeignlanguage{en}{Handbook of {{Model
  Checking}}}}.\hskip 1em plus 0.5em minus 0.4em\relax {Springer International
  Publishing}, 2018, pp. 277--303.

\bibitem{biere_BoundedModelChecking_2003}
A.~Biere, A.~Cimatti, E.~M. Clarke, O.~Strichman, and Y.~Zhu,
  ``\BIBforeignlanguage{en}{Bounded {{Model Checking}}},'' in
  \emph{\BIBforeignlanguage{en}{Advances in {{Computers}}}}.\hskip 1em plus
  0.5em minus 0.4em\relax {Elsevier}, 2003, vol.~58, pp. 117--148.

\bibitem{demoura_Z3EfficientSMT_2008}
L.~{de Moura} and N.~Bj{\o}rner, ``\BIBforeignlanguage{en}{Z3: {{An Efficient
  SMT Solver}}},'' in \emph{\BIBforeignlanguage{en}{Tools and {{Algorithms}}
  for the {{Construction}} and {{Analysis}} of {{Systems}}}}.\hskip 1em plus
  0.5em minus 0.4em\relax {Springer Berlin Heidelberg}, 2008, pp. 337--340.

\bibitem{dutertre_YICESSMTSolver_2006}
B.~Dutertre and L.~M. de~Moura, ``The {{YICES SMT Solver}},'' 2006.

\bibitem{barrett_CVC4_2011}
C.~Barrett, C.~L. Conway, M.~Deters, L.~Hadarean, D.~Jovanovi{\'c}, T.~King,
  A.~Reynolds, and C.~Tinelli, ``\BIBforeignlanguage{en}{{{CVC4}}},'' in
  \emph{\BIBforeignlanguage{en}{Computer {{Aided Verification}}}}.\hskip 1em
  plus 0.5em minus 0.4em\relax {Springer Berlin Heidelberg}, 2011, pp.
  171--177.

\bibitem{he_EfficientSymbolicReactive_2019}
K.~He, A.~M. Wells, L.~E. Kavraki, and M.~Y. Vardi,
  ``\BIBforeignlanguage{en}{Efficient {{Symbolic Reactive Synthesis}} for
  {{Finite}}-{{Horizon Tasks}}},'' p.~7, 2019.

\bibitem{huang_ControllerSynthesisLinear_2016}
Z.~Huang, Y.~Wang, S.~Mitra, and G.~E. Dullerud, ``Controller {{Synthesis}} for
  {{Linear Dynamical Systems}} with {{Adversaries}},'' in \emph{3rd {{ACM
  Symposium}} and {{Bootcamp}} on the {{Science}} of {{Security}}
  ({{HoTSoS}})}.\hskip 1em plus 0.5em minus 0.4em\relax {ACM}, 2016, pp.
  53--62.

\bibitem{huang_ControllerSynthesisInductive_2015}
Z.~Huang, Y.~Wang, S.~Mitra, G.~E. Dullerud, and S.~Chaudhuri, ``Controller
  {{Synthesis}} with {{Inductive Proofs}} for {{Piecewise Linear Systems}}:
  {{An SMT}}-{{Based Algorithm}},'' in \emph{54th {{IEEE Conference}} on
  {{Decision}} and {{Control}} ({{CDC}})}, 2015, pp. 7434--7439.

\bibitem{pnueli1977temporal}
A.~Pnueli, ``The temporal logic of programs,'' in \emph{18th Annual Symposium
  on Foundations of Computer Science (sfcs 1977)}.\hskip 1em plus 0.5em minus
  0.4em\relax IEEE, 1977, pp. 46--57.

\bibitem{gedik_LocationPrivacyMobile_2005}
B.~Gedik and {Ling Liu}, ``Location {{Privacy}} in {{Mobile Systems}}: {{A
  Personalized Anonymization Model}},'' in \emph{25th {{IEEE International
  Conference}} on {{Distributed Computing Systems}} ({{ICDCS}}'05)}, 2005, pp.
  620--629.

\bibitem{choudhary_DistributedTrajectoryEstimation_2016}
S.~Choudhary, L.~Carlone, C.~Nieto, J.~Rogers, H.~I. Christensen, and
  F.~Dellaert, ``Distributed trajectory estimation with privacy and
  communication constraints: {{A}} two-stage distributed {{Gauss}}-{{Seidel}}
  approach,'' in \emph{2016 {{IEEE International Conference}} on {{Robotics}}
  and {{Automation}} ({{ICRA}})}, 2016, pp. 5261--5268.

\bibitem{li_CoordinatedMultirobotPlanning_2019}
L.~Li, A.~Bayuelo, L.~Bobadilla, T.~Alam, and D.~A. Shell, ``Coordinated
  multi-robot planning while preserving individual privacy,'' in \emph{2019
  {{International Conference}} on {{Robotics}} and {{Automation}} ({{ICRA}})},
  2019, pp. 2188--2194.

\bibitem{dams_AbstractionAbstractionRefinement_2018}
D.~Dams and O.~Grumberg, ``\BIBforeignlanguage{en}{Abstraction and
  {{Abstraction Refinement}}},'' in \emph{\BIBforeignlanguage{en}{Handbook of
  {{Model Checking}}}}.\hskip 1em plus 0.5em minus 0.4em\relax {Springer
  International Publishing}, 2018, pp. 385--419.

\bibitem{holzmann_SpinModelChecker_2008}
G.~J. Holzmann, \emph{\BIBforeignlanguage{en}{The Spin Model Checker: Primer
  and Reference Manual}}, 4th~ed.\hskip 1em plus 0.5em minus 0.4em\relax
  {Addison-Wesley}, 2008.

\bibitem{kwiatkowska_PRISMVerificationProbabilistic_2011}
M.~Kwiatkowska, G.~Norman, and D.~Parker, ``\BIBforeignlanguage{en}{{{PRISM}}
  4.0: {{Verification}} of {{Probabilistic Real}}-{{Time Systems}}},'' in
  \emph{\BIBforeignlanguage{en}{Computer {{Aided Verification}}}}.\hskip 1em
  plus 0.5em minus 0.4em\relax {Springer Berlin Heidelberg}, 2011, vol. 6806,
  pp. 585--591.

\bibitem{saboori_VerificationInitialstateOpacity_2013}
A.~Saboori and C.~N. Hadjicostis, ``Verification of initial-state opacity in
  security applications of discrete event systems,'' \emph{Information
  Sciences}, vol. 246, pp. 115--132, 2013.

\bibitem{yin_NewApproachSynthesizing_2015}
X.~Yin and S.~Lafortune, ``\BIBforeignlanguage{en}{A new approach for
  synthesizing opacity-enforcing supervisors for partially-observed
  discrete-event systems}.''\hskip 1em plus 0.5em minus 0.4em\relax {IEEE},
  2015, pp. 377--383.

\bibitem{mphyper_HttpsGitlabOit_2019}
\BIBentryALTinterwordspacing
CPSL@Duke, ``{Motion Planning using HyperProperties},'' 2019. [Online].
  Available: \url{https://gitlab.oit.duke.edu/cpsl/mp_hyper}
\BIBentrySTDinterwordspacing

\end{thebibliography}

\end{document}